# Trajectory Adaptive Prediction for Moving Objects in Uncertain Environment

Hu Jin

College of Computer Science, University of South China,421001,Hengyang Chnia

Hzz1988@usc.edu.cn

**Abstract.**

The existing methods for trajectory prediction are difficult to describe trajectory of moving objects in complex and uncertain environment accurately. In order to solve this problem, this paper proposes an adaptive trajectory prediction method for moving objects based on variation Gaussian mixture model (VGMM) in dynamic environment (ESATP). Firstly, based on the traditional mixture Gaussian model, we use the approximate variational Bayesian inference method to process the mixture Gaussian distribution in model training procedure. Secondly, variational Bayesian expectation maximization iterative is used to learn the model parameters and prior information is used to get a more precise prediction model. Finally, for the input trajectories, parameter adaptive selection algorithm is used automatically to adjust the combination of parameters. Experiment results perform that the ESATP method in the experiment showed high predictive accuracy, and maintain a high time efficiency. This model can be used in products of mobile vehicle positioning.

## 1 Introduction

The rapid development of smart city construction has greatly promoted the vehicular ad hoc network, VANET) and the wide application of related technologies. The Internet of Vehicles (iot) is a huge interactive network composed of information such as vehicle location, speed and route. The birth and development of the Internet of Vehicles is of great help to active vehicle safety, intelligent traffic management, urban life services, emergency rescue and other applications, which are mostly inseparable from the analysis and prediction of location data. The uncertainty of the actual traffic environment brings great challenges to the acquisition of the real-time spatial position of the moving object. If the movement environment deteriorates suddenly, the communication signal becomes weak and the global positioning system (GPS) becomes weak system, GPS) signal will be interrupted for a short time, so that GPS can not obtain real-time position information. This will cause the loss of vehicle track data, affect the transmission of data in the Internet of vehicles, information cannot be shared between vehicles, and eventually lead to serious traffic accidents. In this context, the main scientific problem to be solved in this paper is: how to use the vehicle historical track data to make real-time track position prediction in the dynamic environment of the Internet of vehicles, when the location of mobile network nodes (mobile vehicles) is complex and changeable, the network topology changes frequently, and the difficulty of obtaining traffic information increases, so as to provide the basic data required for decision-making of the intelligent traffic control system.

The trajectory prediction algorithm or model of moving objects has attracted the attention and research of many experts and scholars at home and abroad in recent decades, and has achieved many achievements, and has been widely used in many fields such as transportation, military, medical and daily life services [1-5]. For example, by analyzing the original GPS historical location data, the traffic jam area can be predicted [1]. Measure individual trajectory information entropy to predict human dynamic behavior [2]. MyWay establishes a prediction system, using the single user behavior model, the collective user behavior model and the combination of individual and collective behavior model respectively to predict the next location of mobile users [3]. The advantage of this method lies in that it only needs to share individual mobile configuration files to represent user actions in a concise way, rather than using the original trajectory data to reveal the detailed movements of users. Study uncertain destinations according to the principles of opportunity path and destination, and provide an ideal diversion path to the main destination [4]. Through the prediction and analysis of historical track data, the prediction framework is designed to improve pedestrian safety [5]. At present, the uncertain trajectory prediction methods of moving objects are mainly divided into model prediction methods, linear model prediction methods and nonlinear model prediction methods. The model-based prediction method is to find out the typical trajectory patterns through frequent pattern mining, and then make prediction based on the trajectory patterns [6-9]. Deb[6] et al. proposed to find out the movement pattern matching the most frequent path recorded in the database by mining the track frequent pattern, and then infer the next intersection most likely to be reached. This method does not consider the influence of noise track data, so the prediction accuracy is not high. Commonly used prediction methods based on linear models mainly include Bayesian model prediction method [10-11], Markov model prediction method [12-14] or hidden Markov model prediction method [15-16], and multi-order Markov model prediction method [17-18]. Schreier et al. [11] proposed an operation-based long-term trajectory prediction and risk degree assessment scheme for driver assistance systems in Bayesian networks, which effectively avoided vehicle collisions. Nonlinear model prediction method refers to describing the motion trajectory of moving objects through mathematical formulas, among which Gaussian model [19-21] is a commonly used nonlinear prediction model. Elnaz et al. [19] combined linear regression analysis and Gaussian process regression analysis to predict the long-term trajectory. It has been proved that the model proposed by this method has few parameters and strong stability, but it is only suitable for small-scale data sets. In view of the complex motion mode, jurgen et al. [20] proposed to combine the Bayesian network and Gaussian mixture model to predict the path that the vehicle will travel. This method has high prediction accuracy and low time cost. In the follow-up work of this paper, the method of using Bayesian network to process the mixed Gaussian model to optimize the model parameters is referred to, and the number of mixed Gaussian components is flexibly adjusted. Some scholars make use of the social relations of trajectory [22] and the spatio-temporal information of trajectory [23-24] to make prediction. Most of the above methods only consider the influence of historical track, movement speed, direction, time, geographical features and other factors on the prediction accuracy, and rarely study the influence of dynamic environment on the prediction results. Aiming at the problems of low accuracy and poor real-time performance of the existing prediction methods in the dynamic environment, this paper proposes an adaptive trajectory prediction method in the dynamic environment. The method proposed in this paper can automatically adjust the number combination of Gaussian components of the Gaussian mixture model to adapt to the current environment by introducing the variable decibel Bayesian approximate reasoning method and the adaptive parameter selection algorithm to the traditional Gaussian mixture model, improve the robustness of the prediction model to the complex environment, maintain high accuracy and low time cost.

Section 2 of this paper introduces the prediction model, Section 3 introduces the environment adaptive trajectory prediction algorithm based on the VGMM model, Section 4 is the simulation experiment evaluation of the proposed prediction model, and finally the full text summary.

## 2 Prediction Model for Moving Objects

The trajectory prediction of moving objects in dynamic environment mainly consists of four steps:

(1) Acquisition of original position data, collecting real-time position information of vehicles through vehicle-mounted mobile terminals; (2) Preprocessing of historical track data, removing noise track and extracting local track data features by simplifying, clustering and segmenting the historical track data; (3) Trajectory data modeling: conduct dimensionality reduction operations on spatio-temporal trajectory data, extract motion modes, and build efficient prediction models; (4) Online trajectory prediction. Based on trajectory feature correlation, the movement trend of moving objects is analyzed and the trajectory is simulated by prediction model.

**2.1 Trajectory data preprocessing**

Since the original data about the location information of moving objects received by GPS and other mobile positioning devices contain a lot of noise and redundant data, which cannot be directly used for trajectory prediction, it is necessary to pre-process the position data before using the variational Gaussian mixture model for trajectory data modeling. Firstly, the K-Means clustering method is used for preliminary clustering of trajectory data, and the data is partitioned. Considering that K-Means clustering method has poor clustering effect on irregular shape sets and is sensitive to noise data, density-based clustering method has better processing of noise data. Therefore, this section then uses density-based clustering method to further cluster the track data and remove the noise data in the original track data. The density-based clustering method inputs the neighborhood radius (Eps) and the minimum number of domain data (MinPts) to determine the core points, and then divides the data into core points, noise points and edge points according to the distance from the core points, so the noise points are separated. Secondly, the density-based trajectory segmentation algorithm [9] is used to further process trajectory data, extract local trajectory data features, and improve the time efficiency of the algorithm. The specific implementation process of the clustering method proposed in this paper is referred to literature [25]. The density-based trajectory segmentation method is realized by using breadth first search strategy.

**2.2 Model Training**

Due to the existence of typical non-Gaussian noise in the complex traffic environment, the performance of the traditional Gaussian noise model in this complex traffic environment will suffer a great loss, even can not work. At present, Gaussian mixture model is the most widely used model for non-Gaussian noise modeling. However, most estimation methods for Gaussian mixture distribution parameters are insufficient. For example, although the traditional EM algorithm convergence speed is fast, there may be overfitting phenomenon or local optimal solution. The traditional parameter estimation of Gaussian mixture model can not make good use of prior information, so the obtained posterior probability may not be the true maximum. In this paper, a variational Gaussian mixture model is used to model the trajectory data of moving objects. Variational Gaussian mixture model refers to a mixed distribution obtained by applying variational Bayesian inference on the basis of the traditional mixed Gaussian model, namely, the Bayesian processing of the traditional mixed Gaussian model. The influence of non-Gaussian noise on trajectory prediction accuracy can be solved by introducing a hidden variable Z to the traditional Gaussian mixture model, which indicates that each D-dimensional observation sample Xn originates from a specific Gaussian distribution. The Gaussian mixture model is a linear combination of K Gaussian distributions $P(x) = \sum_{k} \varpi_k N(x | \mu_k, \Lambda_k^{-1})$, $(\varpi_k, \mu_k, \Lambda_k)$ are the weight, mean value and precision matrix of the KTH Gaussian member respectively, and the inverse matrix of the precision matrix is the covariance matrix. The probability density function of variational Gaussian mixture model can be expressed as [26]

$$p(Y|Z,\pi,\mu,\Lambda) = \prod_{n=1}^{N}\prod_{k=1}^{K}\{\frac{1}{(2\pi)^{\frac{D}{2}}}|\Lambda_k|^{\frac{1}{2}} \times \exp[-\frac{1}{2}(y_n-\mu_k)^T\Lambda_k(y_n-\mu_k)]\}^{z_{nk}}$$

In the above equation, Z represents the hidden variable, D is the dimension of observation data, and $(\mu_k,\Lambda_k)$ is the mean value matrix and precision matrix of the KTH Gaussian component respectively.

The essence of using VGMM model to model trajectory data is to accurately estimate the parameters of the model. The parameter estimation of non-Gaussian noise model is often estimated by using the variable decibel Bayesian approximate learning method. Variational Bayes is a learning method proposed by introducing variational approximation theory on the basis of traditional Bayesian inference and examination maximum EM iterative estimation algorithm. Also known as variation bayesian Examination maximum (VBEM) algorithm, it effectively combines the advantages of EM iterative algorithm, such as fast convergence speed, simple solving process and Bayesian inference theory using existing prior information. It improves the problem that the traditional Gaussian mixture model uses EM iterative algorithm to calculate the model parameters easily fall into the local optimal solution. Suppose the observation data set X={x1,...... ,xN}, N data are independently and equally distributed, the prior distribution of all parameters is known, and the hidden variables and parameters are respectively defined by Z={z1,...... ,zN} and θ={θ1,...... ,θN}, and the variational reasoning logarithm edge likelihood function can be approximated as [27].

$$\ln p(X) = F(q(Z,\theta)) + KL(q(Z,\theta) \| p(Z,\theta|X))$$

Where

$$F(q(Z,\theta)) = \iint q(Z,\theta)\ln\frac{p(X,Z,\theta)}{q(Z,\theta)}dZd\theta$$

$$KL(q(Z,\theta) \| p(Z,\theta|X)) = \iint q(Z,\theta)\ln\frac{q(Z,\theta)}{p(Z,\theta|X)}dZd\theta$$

The second $KL(q(Z,\theta) \| p(Z,\theta|X)) \geq 0$ term is the KL divergence between $q(Z,\theta)$ and $p(Z,\theta|X)$. To find the minimum value of the log-likelihood function, we need to first calculate the minimum value of the KL divergence. However, it is difficult and complicated to calculate the minimum value of KL divergence. Therefore, only the maximum value of free energy $F(q(Z,\theta))$ can be calculated.

Parameter $\theta = \{\pi,\mu,\Lambda\}$, the prior distribution of mix weight $\pi$ meet Dirichlet distribution $Dir(\pi|\alpha)$, the mean $\mu$ follow Gaussian distribution $N(\mu|m,(\beta\Lambda)^{-1})$, precision $\Lambda$ meet Weast distribution $W(\Lambda|w,v)$. The Parameter $T = \{\alpha,\beta,m,w,v\}$ is hyper-parameter. The conditional probability density function of hidden variable Z and mixed weight π is

$$p(Z|\pi) = \prod_{n=1}^{N}\prod_{k=1}^{K} \pi_k^{z_{nk}}, \quad z_{nk} \text{ is a discrete factor of 0 or 1.}$$

According to the variational learning mean field theory, it can be assumed that the hidden variable Z and the parameters are independent of each other, so the factorization of the joint density function of the approximate solution is $q(Z,\pi,\mu,\Lambda) = q(Z)q(\pi)\prod_{k=1}^{K} q(\mu_k|\Lambda_k)q(\Lambda_k)$. This can be obtained the optimal $q(Z)$, $q(\pi)$, $q(\mu,\Lambda)$ by iterative calculation with EM method. The process of solving the approximate posterior distribution of hidden variable Z is understood as E-step, and the approximate joint posterior distribution of parameters $(\pi,\mu,\Lambda)$ is understood as M-step[28-29]. When sample prior probability $q(Z)$ reaches the maximum value, EM iteration ends.

Expectation Step: E-step

$$q(Z) \cong \exp\{\sum\sum_{z_{nk}} [\frac{1}{2}E(\Lambda)\ln|\Lambda_k| - \frac{D}{2}\ln(2\pi) - \frac{1}{2}E(\Lambda,\mu)$$
$$[(x_n - \mu_k)'\Lambda_k(x_n - \mu_k)] + E(\pi)(\ln\pi_k)]\} = \prod_{n=1}^{N}\prod_{k=1}^{K} r_{nk}^{z_{nk}}$$

Where $E(z_{nk}) = r_{nk}$.

Maximization Step: M-step

$$q(\pi,\mu,\Lambda) \cong \exp[\sum_{n=1}^{N}\sum_{k=1}^{K} E(z_{nk})$$
$$(\ln N(x_n|\mu_k,\Lambda_k^{-1}) + \ln(\pi_k)) + \ln Dir(\pi|\alpha) +$$
$$\sum_{k=1}^{K} \ln(N(\pi_k|m_0,(\beta_0\Lambda_k)^{-1})W(\Lambda_k|w_0,v_0))]$$

In the formula, the parameter m0 represents the prior mean value, the gradient coefficient of the β0 covariance matrix, W0 is the prior accuracy value, and v0 is the initial value of the freedom of the Wesshot distribution.

Updata parameters

$$\alpha_k = \alpha_0 + N_k$$
$$m_k = \frac{1}{\beta_0 + N_k}(\beta_0 m_0 + N_k \bar{x}_k)$$
$$\beta_k = \beta_0 + N_k, v_k = v_0 + N_k$$
$$w_k^{-1} = w_0^{-1} + N_k S_k + \frac{\beta_0 N_k}{\beta_k}(\bar{x}_k - m)(\bar{x}_k - m)^T$$

The observation data set obeys the following approximate Gaussian mixture distribution when the Gaussian mixture model is processed by the variable decibel Bayesian approximate inference method

$$p(x) \cong \frac{1}{\alpha}\sum_{k=1}^{K} \alpha_k St(x|m_k,\eta_k,v_k+1-d)$$

In the above equation, $m_k$ and $\eta_k$ are the mean value and precision of the $K_{th}$ component respectively. The calculation method of $\eta_k$ is as follows

$$\eta_k = \frac{(v_k + 1 - d)}{1 + \beta_k} w_k$$

Assume that the training data set is Dtrain=(xh,xf), in which the input data is xh, the output data is xf, the test data is Dtest=(x*h,x*f), the input test data is x*h, and the predicted output is x*f. According to formula (5), the joint probability density function and edge probability density function of [xf, x*f] can be obtained, and then the conditional probability density function can be obtained by using probability and statistics theory, which follows the following VGMM model

$$p(x_f | x_h) = \frac{1}{\alpha} \sum_{k=1}^{K} \hat{\alpha}_k St(x_f | x_h, \hat{m}_k, \hat{\eta}_k, \hat{v}_k + d)$$

Where

$$\hat{v}_k = v_k + 1 - d$$

$$\hat{\alpha}_k = \frac{\alpha_k St(x_h | m_{k,x_h}, \eta_{k,x_h}, \hat{v}_k)}{\sum_{j=1}^{k} \alpha_j St(x_h | m_{k,x_h}, \eta_{k,x_h}, \hat{v}_k)}$$

$$\hat{m}_k = m_{k,x_f} + \sum_{k,x_f,x_h} \sum_{k,x_h,x_h}^{-1} (x_h - m_{k,x_h})$$

So, the regression function of x*f with respect to xf, that is, the predicted value of xf is

$$x_f^* = E(x_f^* | x_f) = \sum_{k=1}^{K} \hat{\alpha}_k E(x_{k,f}^* | x_{k,f})$$

Formula (8) is the trajectory variational Gaussian mixture regression model. The basic idea is as follows: First, Formula (7) is used to model the trajectory data by using probability density function, and the training trajectory data is analyzed and processed by density clustering and segmentation methods. Then, the parameters of the Gaussian mixture model were calculated iteratively by using the variable decibel Bayesian approximate inference, and the regression functions of K Gaussian components were obtained according to the conditional distribution of the data conforming to the variational Gaussian mixture distribution. Finally, Formula (8) was used to complete the trajectory regression prediction by weighting the regression function.

This section presents the system model used for the location-based approach for privacy protection and service evaluation in an edge computing environment, describing the data processing module and the result evaluation module accordingly.

## 3 Environment Adaptive Trajectory Prediction Algorithm based on VGMM
### 3.1 Principle of Operation

Based on the variational Gaussian mixture model, the environmental adaptive trajectory prediction algorithm proposed in this paper abstracts the complex trajectory patterns corresponding to the VGMM model, and uses the variational query algorithm to solve the estimation problems of the weight, mean value and accuracy of the GMM model to complete the trajectory prediction. In the process of using VGMM model to model the complex trajectory motion mode, it is necessary to firstly segment the massive position data and divide the trajectory used for training into different segments, using {Sii=1,2,3,......,N} represents. Meanwhile, grid sequence is used to represent the trajectory and simplify the trajectory, using {Oi,i=1,2,3,......,K} represents. Then, the model is trained and learned, and the approximate posterior probability distribution is obtained iteratively by the variational Bayesian expectation maximization method, and the conditional probability density function of the historical trajectory and the future trajectory is designed. Finally, conditional probability density function is used to design the regression prediction function of future trajectory about historical trajectory. In the trajectory prediction stage, input the test trajectory data set and calculate the predicted value using the regression prediction function obtained in the process of model training.

## 3.1 Adaptive Trajectory Prediction based on VGMM Environment

ETAP-VGMM consists of two stages: 1) In the training stage, the historical track data of the moving object processed by the parameter adaptive selection algorithm is taken as the training data set to discover the typical motion pattern of the track, and then the efficient prediction model based on the variational Gaussian mixture model is established. 2) In the prediction stage, the prediction model obtained in the training stage is used to predict the trajectory of the moving object in a short period of time in the future.

## 4 Simulation

In this section, a series of experiments will be conducted using the location data of vehicles in real traffic scenes to evaluate the performance of the proposed method. In this section, the data sources and experimental environment used in the experiment are firstly introduced. Second, performance is measured by evaluation criteria; Finally, show the results.

The average prediction error) : defined as the geometric space error between the predicted track point and the real track point. Suppose that the predicted track point (x´i,y ´i), the actual track point (xi,yi), and the average prediction error of k-step are calculated by the following formula

$$RMSE = \frac{\sum_{i=1}^{k} \sqrt{(x_i - x_i^{'})^2 + (y_i - y_i^{'})^2}}{k}$$

Accuracy: defined as the frequency of occurrence of a real track point in the predicted track point set. If the predicted track point is a real track point, p(l) is 1; otherwise, it is 0.

$$Accu_T = \frac{1}{T} \sum_{l \in T} p(l)$$

### 4.1 Performance Evaluation

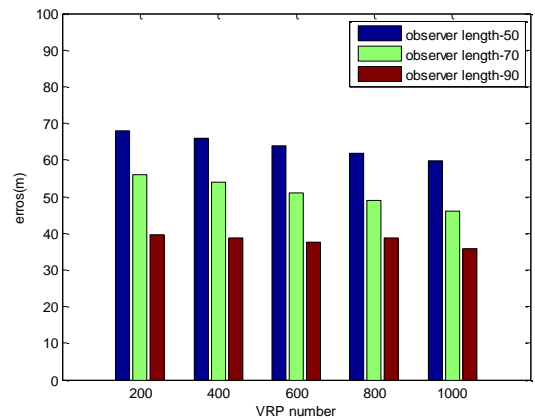

Fig.3 Prediction error under different observable length of input trajectories.

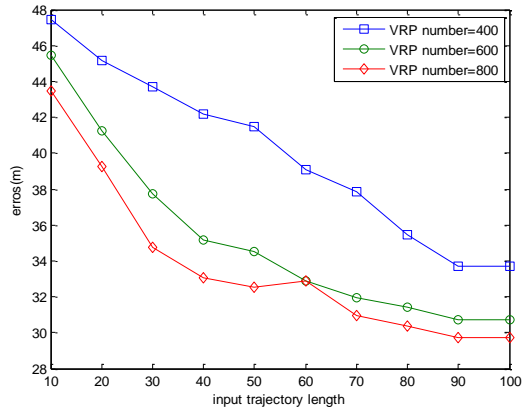

Fig .4 Average precision under different observable length of input trajectories

## 5 Conclusion

In this paper, it is difficult to accurately describe the trajectory of the target in the dynamic environment and there are non-Gaussian noises in the trajectory data, so an adaptive trajectory prediction method based on variational Gaussian mixture model is proposed. In order to predict the trajectory position of moving objects in dynamic environment, clustering method and trajectory segmentation method are firstly used to optimize the trajectory source data. Secondly, the model parameters are estimated by variational Bayesian inference-expectation maximization algorithm to obtain better model parameters, and the parameter combination is adjusted by parameter adaptive selection algorithm, so that the prediction model can adapt to different environments. The experimental results show that the proposed algorithm has high prediction accuracy while maintaining low time cost.